\documentclass[pdflatex,sn-mathphys-num]{sn-jnl}
\usepackage{graphicx}%
\usepackage{multirow}%
\usepackage{amsmath,amssymb,amsfonts}%
\usepackage{amsthm}%
\usepackage{mathrsfs}%
\usepackage[title]{appendix}%
\usepackage{cancel}%
\usepackage{textcomp}%
\usepackage{manyfoot}%
\usepackage{booktabs}%
\usepackage{url}
\usepackage{colortbl}
\usepackage{subcaption}
\usepackage{algorithm}%
\usepackage{algorithmicx}%
\usepackage{algpseudocode}%
\usepackage{listings}%

\usepackage{comment}%
\usepackage[table]{xcolor}%

\theoremstyle{thmstyleone}%
\theoremstyle{thmstyletwo}%
\theoremstyle{thmstylethree}%
\definecolor{orange}{RGB}{250,180,150}
\definecolor{red}{RGB}{255, 138, 138} 
\definecolor{green}{RGB}{138, 255, 158} 
\definecolor{blue}{RGB}{134, 125, 255} 
\definecolor{gray}{gray}{0.85}

\begin{document}

\title[Article Title]{Embodied Tactile Perception of Soft Objects Properties}

\author[1]{\fnm{Anirvan} \sur{Dutta}}\email{a.dutta22@imperial.ac.uk}
\author[1]{\fnm{Alexis} \sur{WM Devillard}}\email{a.devillard20@imperial.ac.uk}
\author[1]{\fnm{Zhihuan} \sur{Zhang}}\email{zhihuazhang003@gmail.com}
\author[2]{\fnm{Xiaoxiao} \sur{Cheng}}\email{xiaoxiao.cheng@manchester.ac.uk}
\author*[1]{\fnm{Etienne} \sur{Burdet}}\email{e.burdet@imperial.ac.uk}

\affil[1]{\orgdiv{Department of Bioengineering}, \orgname{ Imperial College of Science, Technology and Medicine}, \orgaddress{\city{London}, \country{United Kingdom}}}
\affil[2]{\orgdiv{Department of Electrical \& Electronic Engineering}, \orgname{The University of Manchester}, \orgaddress{\country{United Kingdom}}}

\abstract{To enable robots to develop human-like fine manipulation, it is essential to understand how mechanical compliance, multi-modal sensing, and purposeful interaction jointly shape tactile perception. In this study, we use a dedicated \textit{modular e-Skin} with tunable mechanical compliance and multi-modal sensing (normal, shear forces and vibrations) to systematically investigate how sensing embodiment and interaction strategies influence robotic perception of objects. Leveraging a curated set of soft 
\textit{wave objects} with controlled viscoelastic and surface properties, we explore a rich set of \textit{palpation primitives}-pressing, precession, sliding that vary indentation depth, frequency, and directionality. In addition, we propose the \textit{latent filter}, an unsupervised, action-conditioned deep state-space model of the sophisticated interaction dynamics and infer causal mechanical properties into a structured latent space. This provides generalizable and in-depth interpretable representation of how embodiment and interaction determine and influence perception. Our investigation demonstrates that multi-modal sensing outperforms uni-modal sensing. It highlights a nuanced interaction between the environment and mechanical properties of e-Skin, which should be examined alongside the interaction by incorporating temporal dynamics.}

%significantly expanding prior studies.}

\maketitle

\section{Introduction}\label{sec:intro}
Humans exhibit a remarkable ability for perceiving and manipulating various objects by performing exploratory actions like squeezing, palpating, and probing. The underlying sensory abilities are largely attributed to the multilayered, adaptive structure of human skin, populated with dense arrays of specialized mechanoreceptors \cite{rawlings2012innovation}. This sensorised skin, together with purposefully deployed exploration strategies, enable the extraction of rich, high-fidelity information during dynamic interactions, where both the skin and the contacted object deform \cite{hayward2011there}. However, the specific contribution of the skin’s mechanical properties and multi-modal sensing or \textit{embodiment}\footnote{Here, \textit{embodiment} refers to how the physical body and its sensory apparatus shape perception and cognition, rather than the notion of agency and self-perception~\cite{zopf2018revisiting}.} to tactile perception remains poorly understood \cite{hayward2008haptic}.

\begin{figure}[!h]
    \centering
    \includegraphics[width=0.75\textwidth]{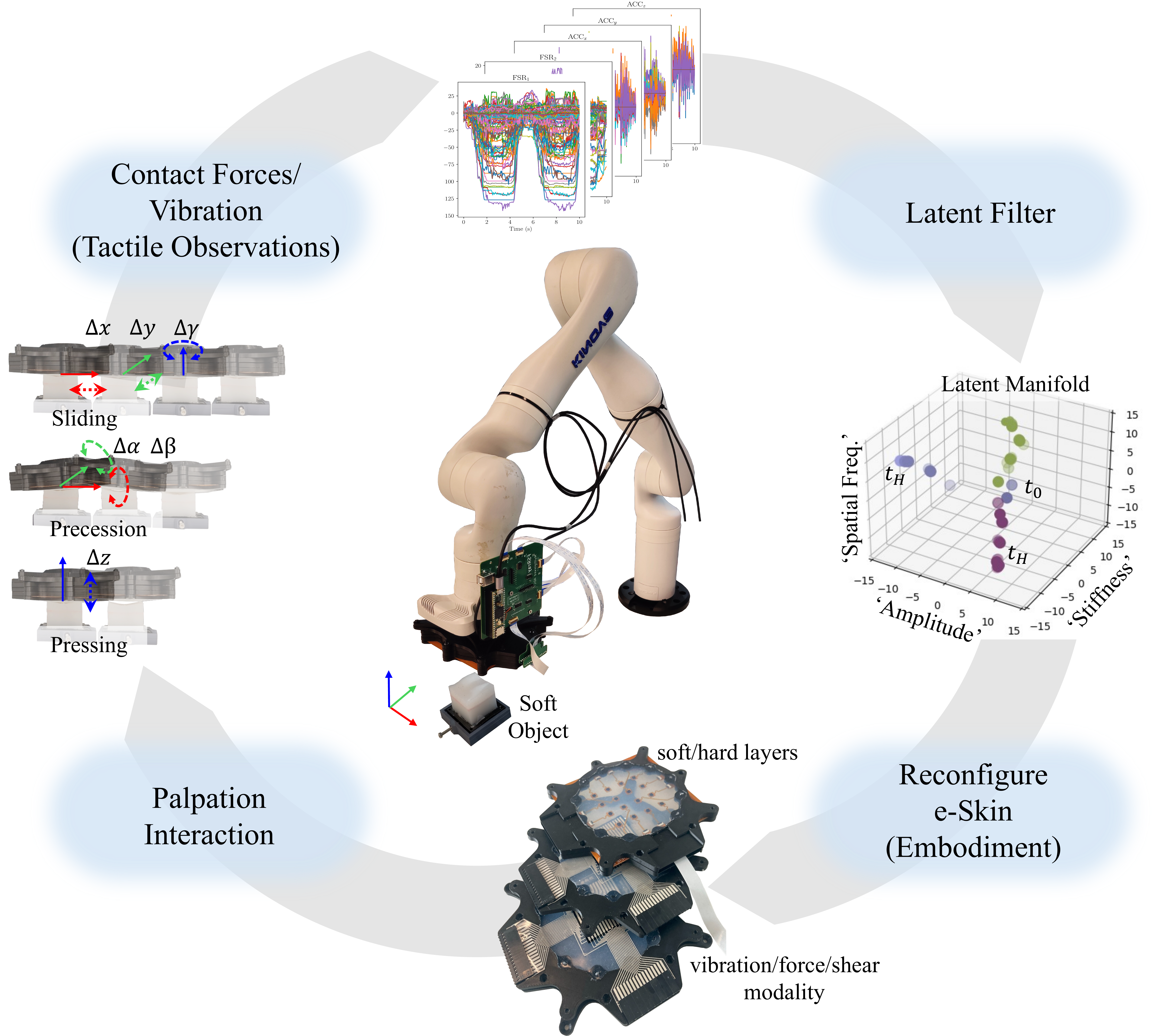}
    \caption{Interactive setup to investigate how skin sensing and embodiment influence tactile sensing.}
    \label{fig:problem_setup}
\end{figure}

Understanding the interplay in embodiment and perception is crucial for robotic systems equipped with tactile sensing to achieve perceptual capabilities of the quality of biological counterparts. Studies by Costi et al. \cite{costi2022environment, costi2022magneto} show that adding a soft outer layer to a force sensing layer improves contact fidelity, but dampens high-frequency signals important for sensing surface texture—effectively acting as a mechanical low-pass filter. Follow-up work has explored how tuning this filter improves performance for specific object categories \cite{hughes2021online}, with softer skins better discriminating compliant materials and stiffer ones suited for rigid objects. Yet, these studies mostly focus on uniform outer layers and did not investigate the influence of multimodal sensing and a multilayered mechanical structure. In this article, we systematically investigate embodied tactile perception by considering the sensory and mechanical aspects of an artificial skin, systematic variations in the environmental surface and mechanical properties, and a repertoire of palpation primitives, as illustrated in Fig.\,\ref{fig:problem_setup}. To this end, we developed a modular \textit{electronic skin} (e-Skin) \cite{alexiswhc25} (see Results~\ref{subsec:eskin}), which allows for controlled modulation of mechanical properties and sensing modalities—enabling systematic exploration of their influence on perception. 

Previous studies have largely relied on simplified environments, such as biological tissue phantoms \cite{qiu2022soft, scimeca_action_2022, herzig_variable_2018} or homogeneous viscoelastic materials with basic geometries \cite{sun_recognition_2023, costi2022environment, costi2022magneto}. While valuable, these configurations do not capture the diversity of naturally occurring soft objects, which often exhibit complex combinations of bulk mechanical properties and surface textures that jointly shape the perception of softness \cite{cavdan2021task, steer2023feel}. To emulate this variability, we curated a foundational dataset of soft objects - \textit{wave objects} (see Results~\ref{subsec:object}) featuring controlled variations in bulk compliance and surface characteristics, enabling a comprehensive evaluation of embodied tactile perception.

In addition, the nature of exploratory interaction is critical to tactile perception. Human studies have shown that individuals adapt their exploration strategies—from simple probing to complex palpation—depending on the perceptual objective \cite{visell_vibrotactile_2014}. In contrast, robotic studies often rely on fixed, static probing motions \cite{amin_embedded_2023}, overlooking how interaction parameters such as indentation depth, speed, and frequency affect perception. Moreover, sensory processing typically neglects temporal interaction dynamics, relying instead on static analyses that fail to capture how perception evolves during interaction. Without systematic investigation of how dynamic, action-dependent exploration shapes tactile signals \cite{bergmann_tiest_tactual_2010}, it remains difficult to develop a generalizable understanding of how embodiment and interaction jointly influence perception across diverse soft objects. To address this, we employ a repertoire of palpation-based interactions—including pressing, precession, and sliding, inspired by human soft object exploration \cite{lezkan2018active} to elicit rich tactile responses during contact (see Methods~\ref{subsec:interaction}), enabling precise identification of salient mechanical properties while accounting for interaction dynamics \cite{dense_phys}.

Beyond these limitations in sensing and interaction, a central challenge lies in modeling soft sensor-object interaction. Most prior approaches are discriminative, relying on handcrafted statistical features \cite{scimeca_structuring_2020, hui2014evaluating}, dimensionality reduction, or sophisticated classifiers to detect stiffness levels \cite{konstantinova_palpation_2017}. Some works have applied techniques such as functional principal component analysis \cite{wang_tactual_2022} or used kinesthetic sensing alone \cite{xu_exploring_2019, bassal_hardness_2024}. While deep learning approaches—using visuo-haptic CNNs \cite{caldiran_visuo-haptic_2019, huang_variable_2022, sun_recognition_2023} or recurrent networks for temporal tactile signals \cite{bednarek_gaining_2021, nonaka_soft_2023}—have shown improved performance, they are typically restricted to predefined object categories and lack physically grounded causal inference. Moreover, their performance is often highly sensitive to the specific soft object set used during training and evaluation, highlighting the need for principled, regression-based models that support robust and unbiased analysis. Although regression-based approaches such as Contact Area Spread Rate (CASR) for bulk viscosity estimation \cite{bicchi_haptic_2000} or spring-damper models with Kalman filtering \cite{uttayopas2023object} offer an alternative, they often depend on strong simplifying assumptions (e.g., material homogeneity or constrained motions) and struggle to generalize across diverse object types. These methods are further limited by their reliance on low-dimensional kinesthetic signals and by their exclusion of high-dimensional tactile information. Standardized metrics for soft object characterization (e.g., stiffness, loss tangent, Shore hardness) are similarly constrained to quasi-static, uni-modal interaction paradigms \cite{estermann_quantifying_2020}. To overcome these challenges, we introduce an unsupervised, action-conditioned variational inference approach based on deep state-space modeling - \textit{Latent Filter} (see Results~\ref{subsec:latentfilter}). This model captures the temporal dynamics of sensor-object interactions and learns a compact latent representation that encodes the underlying mechanical properties of soft objects, enabling generalizable and physically meaningful inference.

\section{Results}
\label{sec:results}
This section presents the outcomes of our study using the proposed framework. We first describe the modular \textit{e‑Skin}, which enables systematic variation of mechanical compliance and sensing modalities, and the curated set of soft objects with controlled viscoelastic and geometric properties. We then introduce the proposed \textit{Latent Filter}—a deep, unsupervised, action‑conditioned state‑space model designed to infer intrinsic soft object properties from temporal tactile observations. Leveraging this framework, we conduct a series of experiments to analyze how sensor embodiment, sensing modalities, and interaction strategies jointly influence the perception of soft object properties.

\begin{figure}[!tbh]
    \centering
    \includegraphics[width=0.85\textwidth]{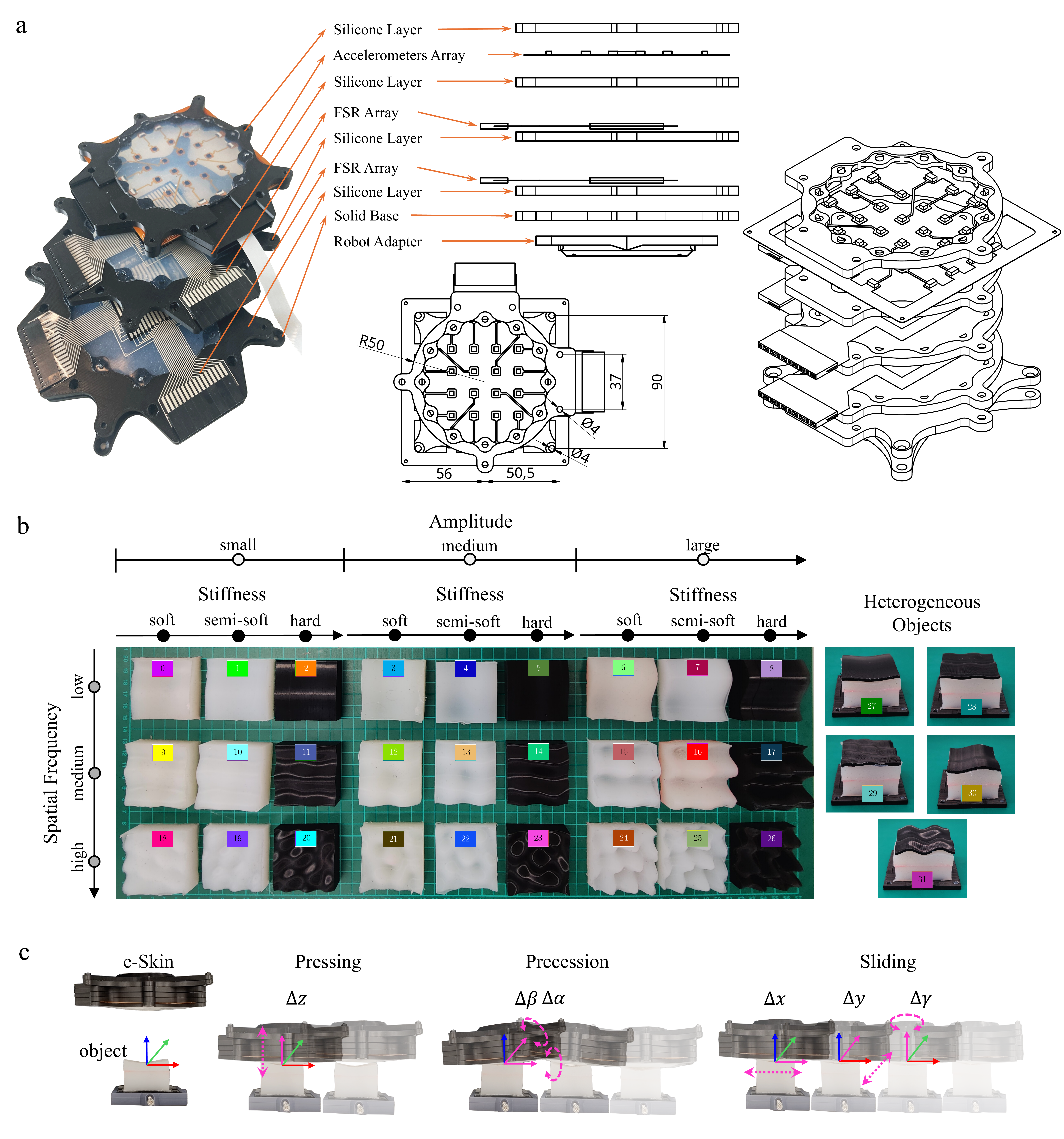}
    \caption{E-skin, object set and movements primitives to investigate tactile perception for robotics. a) Diagram of the modular e-Skin \cite{alexiswhc25} used in the current paper's study. b) \textit{Wave objects} with controlled surface and stiffness properties. Objects 0–26 are homogeneous in material, objects 27–31 are heterogeneous, incorporating a thin stiff top layer. c) Robotic palpation primitives designed to elicit rich tactile responses. Active motion axes are shown in pink.}
    \label{fig:methodsfig}
\end{figure}

\subsection{e-Skin Tactile Sensor}
\label{subsec:eskin}
The modular e-Skin developed for this study features a layered structure, with each layer composed of materials and sensors tailored to control mechanical properties and sensing capabilities. This modularity enables the replacement or reconfiguration of individual layers to suit different tasks and environments. In the configuration used here (Fig.\,\ref{fig:methodsfig}a), the e-Skin comprises three sensorised layers embedded within compliant silicone. Two variants were fabricated by varying the silicone material: a soft e-Skin using Smooth-On Ecoflex (Shore 00-31) and a more rigid e-Skin using Dragonskin (Shore 30A). The outermost layer houses a 4$\times$4 array of multi-nodal 3-axis accelerometers, positioned just beneath a thin silicone layer to minimize damping and detect high-frequency vibrations—analogous to rapidly adapting mechanoreceptors in human skin. To measure normal forces and deformations, two layers of \textit{force-sensing resistors} (FSRs) were embedded, each forming a 16$\times$16 array. FSRs are inherently limited to sensing normal forces, so a compliant inter-layer of silicone was introduced between the two arrays. Shear forces may be indirectly estimated by capturing differential normal force patterns across the FSR layers during shear displacements. This multilayered, multi-nodal configuration enables detailed examination of sensing modalities and mechanical compliance of artificial tactile sensing. 

\subsection{Rich object \textit{wave objects} set}
\label{subsec:object}
To evaluate the e-Skin’s performance in a controlled yet diverse setting of object properties, we designed a set of soft \textit{wave objects} with parametrically controlled surface and stiffness properties (Fig.\,\ref{fig:methodsfig}b). Each object was defined by three parameters: amplitude, spatial frequency (surface texture), and stiffness (bulk compliance). Surface textures were created by applying a band-pass filter to white noise images, centered at spatial frequencies \{10/m,\,30/m,\,50/m\}. This method generated surface profiles with natural variability while retaining control over key parameters. Amplitudes were scaled to one of three levels \{5,10,20\}\,mm. The mirrored surfaces were 3D printed to produce molds for silicone casting. Stiffness variation was achieved using different materials—Ecoflex 00-10, Ecoflex 00-50, and rigid PLA—each with distinct Shore hardness. To introduce heterogeneity, a subset of the objects included a stiff top layer (heterogeneity $\in \{0,1\}$). This dataset captures a wide range of perception conditions relevant to soft object manipulation and provides a foundation for systematic tactile perception evaluation.

\subsection{Deep State-Space Model: Latent Filter}
\label{subsec:latentfilter}
We model the interaction between soft objects and the robot equipped with e-Skin as a discrete nonlinear dynamical system with tactile observations (vibrations, contact forces) $\mathbf{o}_{1:T} = (\mathbf{o}_1, \mathbf{o}_2, .., \mathbf{o}_T), \mathbf{o}_t \in \mathcal{O} \subset \mathbb{R}^{n_o}$ in discrete time steps $t=1, .., T$, and actions $\mathbf{a}_{1:T} = (\mathbf{a}_1, \mathbf{a}_2, .., \mathbf{a}_T), \mathbf{a}_t \in \mathcal{A} \subset \mathbb{R}^{n_a}$. These interactions induce complex spatiotemporal tactile observations that encode information about the object's underlying mechanical properties. An unsupervised deep state-space model is employed to capture interaction dynamics by leveraging time-series and action-conditioned tactile data through low-dimensional latent variables $\mathbf{s}_{1:T} = (\mathbf{s}_{1}, \mathbf{s}_{2}, .., \mathbf{s}_{T})$ with $\mathbf{s}_{t} \in \mathcal{S} \subset \mathbb{R}^{n_s}$ that represent the underlying state of the system.
The objective is to model the joint probability $p(\mathbf{o}_{1:T}, \mathbf{s}_{1:T}|\mathbf{a}_{1:T})$, and to perform variational inference by maximizing the likelihood of observations 
\begin{align}
    p(\mathbf{o}_{1:T}|\mathbf{a}_{1:T}) &= \int  \!\! p(\mathbf{o}_{1:T}, \mathbf{s}_{1:T}|\mathbf{a}_{1:T}) \, d\mathbf{s}_{1:T}
\end{align}
We hypothesize that accurately modeling and inferring the dynamics of the robotic system during its interaction with soft objects will enable the extraction of causal physical factors underlying the process. To support this, we introduce a structural assumption in the latent-space grounded in the concept of observability from classical control theory. Specifically, we distinguish between directly observable components of the latent state-those that can be inferred from a single observation dependent on indirectly observable components-those that require observations accumulated over multiple time steps for accurate estimation. This facilitates analytical computation of the posterior, while avoiding the computational expense of linearizing the observation model. We propose restructuring/partitioning of the latent state-space:
\begin{align}
    p(\mathbf{s}_t) = p(\mathbf{z}_t, \mathbf{y}_t) &= p(\mathbf{z}_t|\mathbf{y}_t) \, p(\mathbf{y}_t)
    \label{eq:factorization}
\end{align}
with directly observable variable $\mathbf{z}_t \subset \mathbb{R}^{n_z}$ and indirectly observable part as $\mathbf{y}_t \subset \mathbb{R}^{n_y}$, $n_s = n_z + n_y$ with updated \textit{Evidence Lower Bound Objective (ELBO)}: (see the supplementary materials, Sec\,1 for detailed derivation):
\begin{align}
    \mathcal{F}_{\textsc{ELBO}}(\theta, \phi)  = \,\mathbb{E}_{q_\phi(.)}\!\left[ \sum_{t=1}^{T} \log p_{\theta}(\mathbf{o}_t|\mathbf{z}_t)\right] 
    - &\sum_{t=2}^{T}{\textsc{KL}}[ q^{filt}(\mathbf{z}_t) || p(\mathbf{z}_{t}|\mathbf{z}_{t-1}, \mathbf{y}_{t}, \mathbf{a}_{t}) \nonumber \\
    -  \sum_{t=2}^{T}{\textsc{KL}}[ &q_\phi(\mathbf{y}_t|.)||p(\mathbf{y}_t| \mathbf{a}_{t})]
    \label{eq:elbo}
\end{align}
\begin{figure}[!tb]
    \centering
    \includegraphics[width=0.85\textwidth]{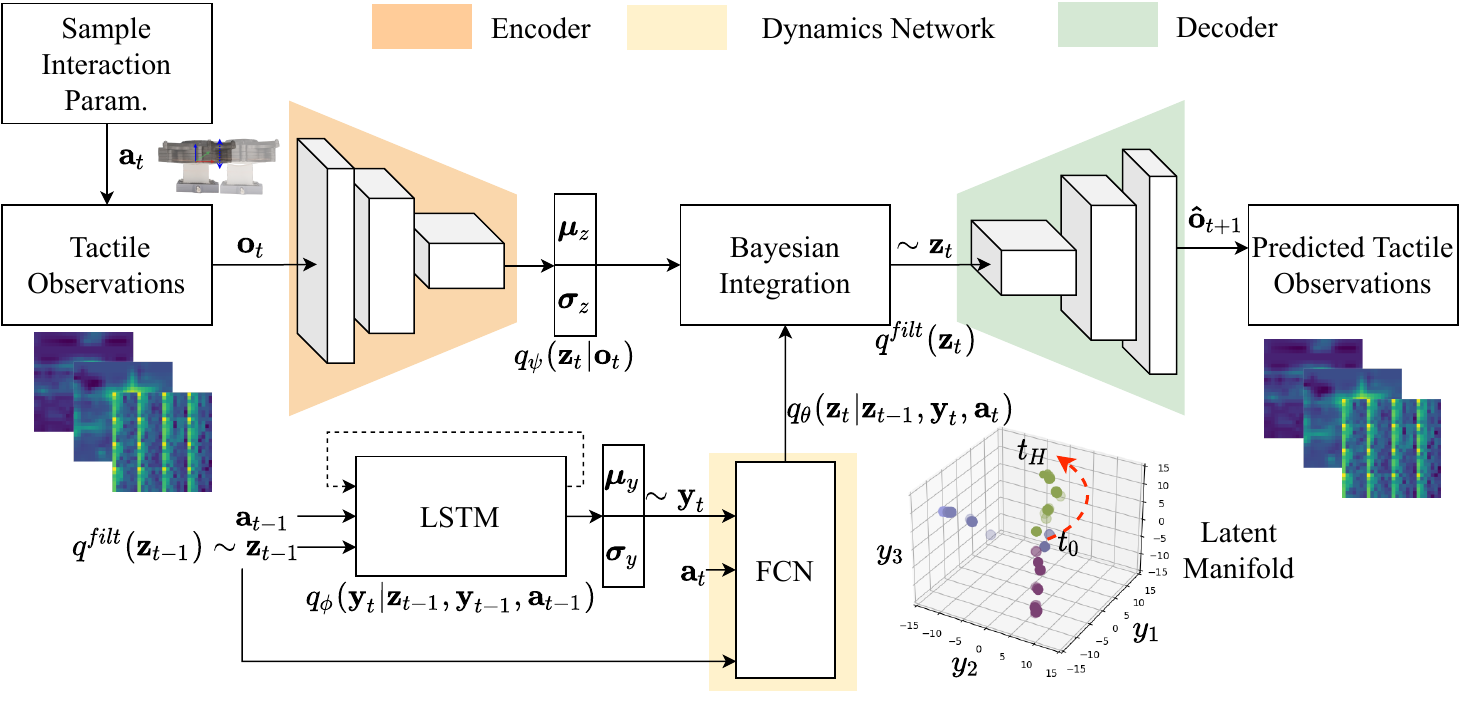}
    \caption{Proposed latent filter architecture. The encoders implement the inverse variational measurement model and are fused via Bayesian integration with the dynamics model, approximated by a fully connected network (FCN). The decoder, mirroring the encoder structure, reconstructs tactile observations. Encoder architectures vary depending on the input modality combination. An LSTM (Long Short-Term Memory) network estimates the indirectly observable latent distribution.}
    \label{fig:framework}
\end{figure}
The architecture of the proposed latent filter is illustrated in Figure~\ref{fig:framework} where we employ the inverse variational measurement model $q_{\psi}(\mathbf{z}_t|\mathbf{o}_t)$ and perform Bayesian integration with the recursive model $q_\theta(\mathbf{z}_t|\mathbf{z}_{t-1}, \mathbf{y}_t, \mathbf{a}_t)$ to compute the filtered variational distribution $q^{filt}(\mathbf{z}_t)$. To approximate the indirectly observable distribution $q_\phi(\mathbf{y}_t|\mathbf{z}_{t-1}, \mathbf{y}_{t-1}, \mathbf{a}_{t-1})$, we utilize an LSTM network. Additionally, we incorporate a learnable hierarchical prior $p(\mathbf{y}_t| \mathbf{a}_t, \mathbb{N})$, where $\mathbb{N}$ represents the soft-object label, to enforce the extraction of causal physical properties while relaxing constraints on time sequence length \cite{latentmatters}. The complete latent filter model is optimized end-to-end using the Evidence Lower Bound (ELBO) objective defined in Eq.~\ref{eq:elbo}.

\subsection{Experiments}
\label{subsec:exp}
We conducted a systematic study examining how sensor embodiment, sensing modality, and interaction strategy jointly influence the perception of soft object properties. Specifically, we investigated the effects of (i) e‑Skin stiffness with soft—Ecoflex vs. hard—DragonSkin, (ii) individual sensing modalities, including the accelerometer layer and the two force-sensing resistor (FSR) layers (top: FSR$_t$, bottom: FSR$_b$), and (iii) three palpation primitives—pressing, precession, and sliding—on the perception of key object properties, namely spatial frequency, amplitude, and stiffness. In addition, we compared two multi‑modal fusion strategies (early and late fusion) to determine how different sensory observations should be integrated to optimize perceptual performance (see Methods~\ref{subsec:dataprocessing}, Figure~\ref{fig:multifusionarc}b). An initial experimental evaluation was performed to identify suitable parameter ranges considering the robot controller frequency and sensor saturation. Within these bounds, 16 uniformly sampled values were selected to perform interaction actions and record tactile observation data. Each trial began with the robot moving to a fixed position, descending until a predefined contact force was reached, and then executing the palpation trajectory for 10\,s. A total of 2560 interaction trajectories were collected, each comprising 7000 samples (sampling frequency: 700\,Hz) for both e-Skin stiffness configurations. Tactile observations included two-layer FSR and accelerometer readings, while action observations (sampling frequency: 100\,Hz) captured the end-effector's position and orientation. This raw observation was processed prior to model training (see Methods~\ref{subsec:dataprocessing}). The following sections present the results obtained across varying e‑Skin configurations and palpation strategies.

\subsection{Analysis of Latent Space}
\label{subsec:static}
\begin{figure}[!h]
    \centering
    \includegraphics[width=0.75\textwidth]{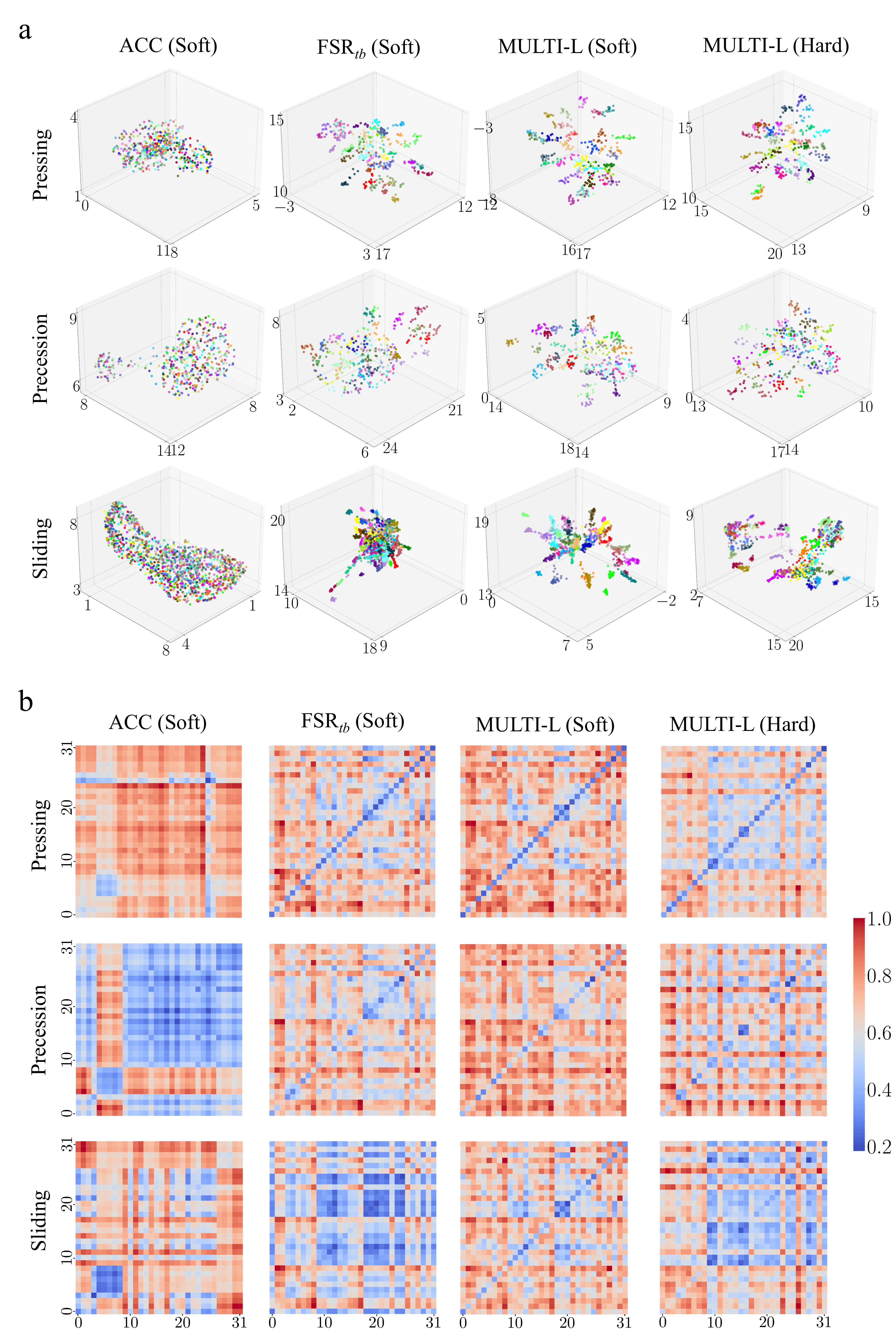}
    \caption{Comparison of latent representations across different tactile sensor modalities, skin stiffness conditions, and interaction primitives. (a) UMAP projections highlight cluster separability, with each unique color signifying a soft object. (b) Distance matrices quantify inter-class similarity in the latent space with colors scaled to min-max values.}
     \label{fig:latentdistance}
\end{figure}

The learned latent features were systematically evaluated. Figure~\ref{fig:latentdistance}\,a presents 3D UMAP \cite{mcinnes2018umap} embeddings of these latent representations (last time step, post-convergence) for different modality configurations: ACC (accelerometer only),  FSR$_{tb}$  (both layers of FSR), MULTI-L (soft) (multi-modal late fusion with soft skin), and MULTI-L (hard) (multi-modal late fusion with hard skin) across the three interaction primitives. 
The UMAP plots reveal that multimodal configurations (MULTI-L), particularly with soft skin, exhibit the most structured and separable clusters, indicating their ability to encode discriminative features. FSR$_{tb}$ alone also produces distinct clusters for many object classes, especially in precession and pressing interactions, underscoring the informativeness of 2-layer contact force information. In contrast, embeddings from the ACC only setting appear more entangled and dispersed, suggesting that transient observations are insufficient for capturing object-specific features. In addition, it also demonstrates that the combination of bulk exploration via pressing and precession provide more discriminative tactile observations than surface level interaction of sliding. As UMAP projections do not preserve relative spatial relationships from the original high-dimensional latent space, both intra-class and inter-class Euclidean distances were computed using object labels to provide a complementary quantitative assessment of latent space separability and confusion among the inferred soft object features. Figure \ref{fig:latentdistance}\,b presents the distances normalized to [0,1] as heatmaps. 

These heatmaps confirm the UMAP observation: MULTI-L (soft), particularly under precession, achieves superior inter-class separability and reduced confusion. The ACC-only modality consistently shows poor discrimination and high confusion specially during precession interaction where consistent stable contact was made. However ACC can detect features by sliding that FSRs alone cannot not distinguish. An intriguing observation is that the MULTI-L (Hard) configuration exhibits increased confusion for object classes 12–26, which are characterized by complex surface features (higher surface spatial frequency). This suggests that the rigid skin configuration may limit the tactile sensor's ability to capture fine surface details critical for class separation. However, euclidean distance metric in such spaces may not be a faithful proxy for true dissimilarity, especially if the latent manifold is non-Euclidean and influenced by hierarchical variational priors. Furthermore, normalization of distances could obscure subtle yet meaningful inter-class distinctions.  

As the synthetic object set is parametrized by known mechanical properties (e.g., stiffness, spatial frequency, amplitude), a more principled analysis was conducted by aligning the latent space with these ground-truth generative factors, using the supervised regression method proposed in \cite{karl2017deep}. This allowed investigation of the time-series evolution of physically meaningful latent features, providing in-depth analysis for this study.

\subsection{Analysis on Multi-Modality of e-Skin}
\label{subsec:reg1}
\begin{figure}[!htb]
    \centering
    \includegraphics[width=0.75\textwidth]{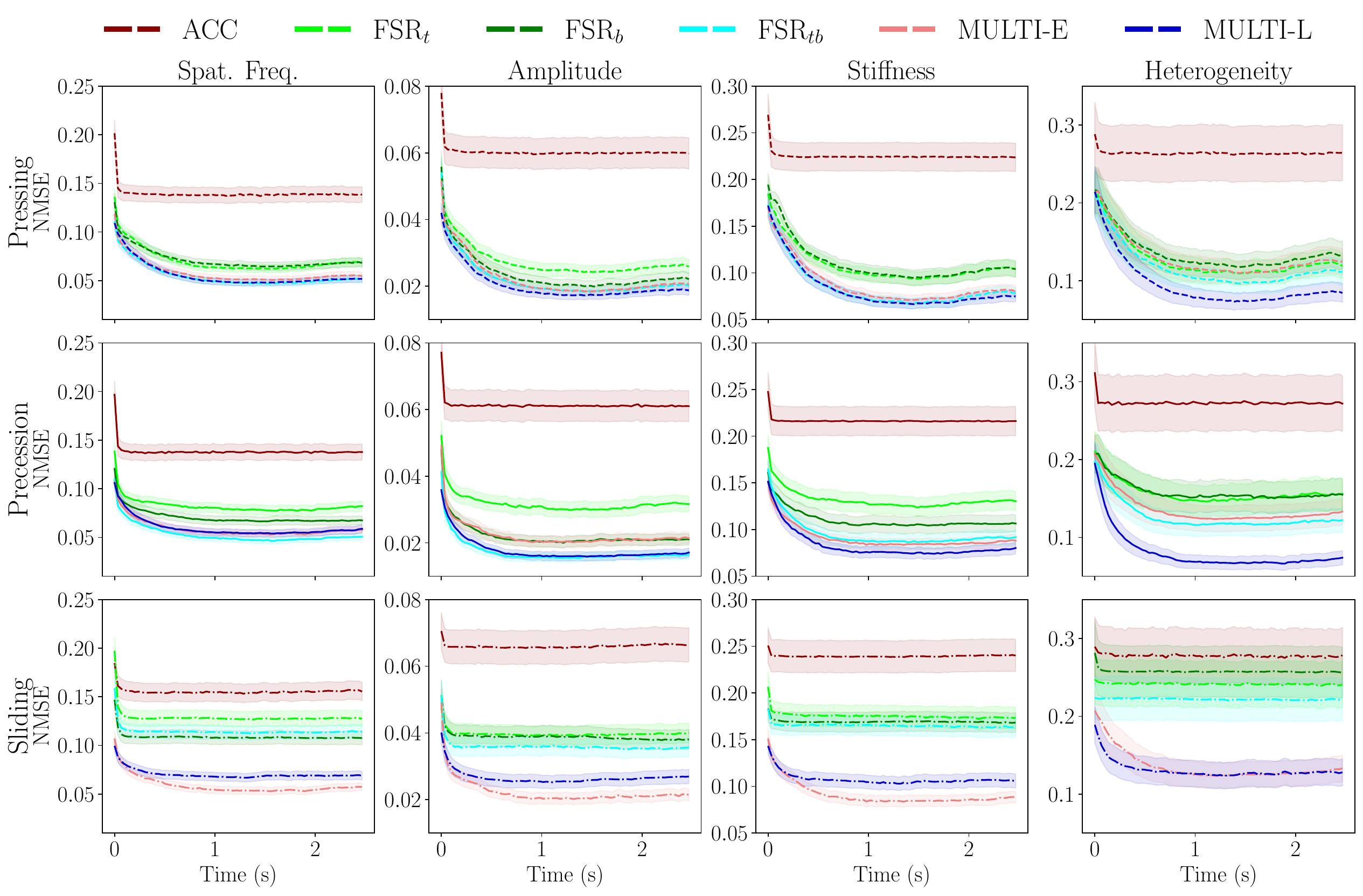}
    \caption{Detailed analysis of the different combination and study of the multi-modality for soft skin. Standard deviation is scaled $\times$10 for visual clarity.}
    \label{fig:multimodal}
\end{figure}

To quantitatively relate the learned latent features to the underlying physical properties, a non-linear kernel ridge regression model \cite{murphy2012krr} was employed. The final segments of each interaction sequence were used to train the regressor, which was then applied to predict the mechanical properties of samples drawn from the evolving latent space. Example plots of two instances of filtering following alignment are depicted in the supplementary material (Fig\,1). To assess alignment performance, the normalized mean squared error (NMSE) was computed between the predicted mechanical properties and ground truth. Figure~\ref{fig:multimodal} presents the temporal evolution of the NMSE, comparing different combinations of modality across the three primitives. The evaluated modalities include: ACC (accelerometer only), FSR$_t$ (first FSR layer), FSR$_b$ (second FSR layer),  FSR$_{tb}$  (both FSR layers), MULTI-E (early fusion of accelerometer and both FSR layers), and MULTI-L (late fusion of accelerometer and both FSR) (see Figure~\ref{fig:multifusionarc}\,b for early and late fusion). For consistency, all of these evaluations were performed using the soft skin. Analyzing the time-series of the prediction error enables assessment of how efficiently each modality combination supports inference. This information is lost if NMSE were averaged over the entire sequence or with other static analysis done in prior works.

The results demonstrate that the late-fusion multimodal approach (MULTI-L) consistently yields the lowest NMSE across all interaction primitives, indicating more robust and discriminative representations. In addition, we can also observe the effect of two-layer FSR in estimating stiffness and heterogeneity property, highlighting the importance of capturing shear force or skin-stretch information. In addition, transient accelerometer signals contribute to faster, albeit less accurate inference, and when combined with two-layer FSR, improves the performance when performing surface level exploration of sliding. Interestingly, although the early fusion scheme (MULTI-E) aligns more closely with the physical nature of sensor co-activation, it results in less consistent performance. This discrepancy reveals a fundamental mismatch in the nature of force and vibrations signals, while force data tends to be smooth and contact-driven, spectrogram of the vibrations signals result in distinct observation distribution indicating that separate encoding pathways are better suited for capturing their respective distribution. These findings highlight not only the need to evaluate multi-modality but also the importance of developing modality-aware tactile encoders capable of disentangling heterogeneous signal sources. Future research could focus on the design of such tactile specific encoders, enabling more efficient and interpretable perception in multi-modal artificial e-Skins.

\subsection{Analysis on Interaction}
\label{subsec:reg2}
The regression approach on the aligned latent values was further applied to investigate the influence of interaction parameters on the perception of soft objects across different components and interaction types. Table~\ref{tab:action_config} lists all the 16 parameters used for the interactions along with their corresponding NMSE values computed over the entire interaction trajectory. We also report (see Supplementary material, Fig\,2) the temporal evolution of the NMSE for best and worst performing parameters, selected to avoid visual and interpretive clutter.

The results reveal that interactions involving moderately higher frequencies ($\sim$0.6\,Hz) and greater contact depths lead to better parameter inference across all interaction types. However, performance deteriorates at the highest tested frequencies, suggesting the existence of an optimal frequency range for effective palpation of soft objects. The influence of interaction parameters is particularly pronounced in precession and sliding, where they significantly affect estimation quality. In contrast, for pressing interactions, parameter variations primarily influence convergence speed rather than final estimation accuracy. Notably, in surface-level sliding, movement amplitude has limited impact, with interaction frequency playing a more dominant role. This analysis provides key insights on the different nature of the palpation interaction for soft-object perception. 

\begin{table*}[!tbh]
\centering
\resizebox{\textwidth}{!}{%
\begin{tabular}{c ccc ccc cccc}
\toprule
\multirow{2}{*}{Action Index} & \multicolumn{3}{c}{Pressing} & \multicolumn{3}{c}{Precession} & \multicolumn{4}{c}{Sliding} \\
\cmidrule(lr){2-4} \cmidrule(lr){5-7} \cmidrule(lr){8-11}
 & $A_{z}\,(mm)$ & $\omega_z$\,(Hz) & NMSE $10^{-2}$ & $A_{\alpha,\beta}$\,(mm) & $\omega_{\alpha,\beta}$\,(Hz) & NMSE $10^{-2}$ & $A_{x,y}$\,(mm) & $A_\gamma(^{\circ})$ & $\omega_{x,y,\gamma}$\,(Hz) & NMSE $10^{-2}$ \\
\midrule
1 & \cellcolor{gray!0!white}2.0 & \cellcolor{gray!0!white}0.2 & \cellcolor{orange!99!white}7.409 & \cellcolor{gray!0!white}3.0 & \cellcolor{gray!0!white}0.2 & \cellcolor{orange!99!white}7.483 & \cellcolor{gray!0!white}2.0 & \cellcolor{gray!0!white}3.0 & \cellcolor{gray!0!white}0.2 & \cellcolor{orange!95!white}10.051 \\
2 & \cellcolor{gray!0!white}2.0 & \cellcolor{gray!33!white}0.4 & \cellcolor{orange!63!white}6.816 & \cellcolor{gray!0!white}3.0 & \cellcolor{gray!33!white}0.4 & \cellcolor{orange!54!white}5.860 & \cellcolor{gray!0!white}2.0 & \cellcolor{gray!0!white}3.0 & \cellcolor{gray!33!white}0.4 & \cellcolor{orange!15!white}7.143 \\
3 & \cellcolor{gray!0!white}2.0 & \cellcolor{gray!66!white}0.6 & \cellcolor{orange!63!white}6.821 & \cellcolor{gray!0!white}3.0 & \cellcolor{gray!66!white}0.6 & \cellcolor{orange!56!white}5.936 & \cellcolor{gray!0!white}2.0 & \cellcolor{gray!0!white}3.0 & \cellcolor{gray!66!white}0.6 & \cellcolor{orange!13!white}7.088 \\
4 & \cellcolor{gray!0!white}2.0 & \cellcolor{gray!99!white}0.8 & \cellcolor{orange!67!white}6.888 & \cellcolor{gray!0!white}3.0 & \cellcolor{gray!99!white}0.8 & \cellcolor{orange!85!white}6.946 & \cellcolor{gray!0!white}2.0 & \cellcolor{gray!0!white}3.0 & \cellcolor{gray!99!white}0.8 & \cellcolor{orange!81!white}9.536 \\
5 & \cellcolor{gray!33!white}4.0 & \cellcolor{gray!0!white}0.2 & \cellcolor{orange!47!white}6.556 & \cellcolor{gray!33!white}4.0 & \cellcolor{gray!0!white}0.2 & \cellcolor{orange!80!white}6.784 & \cellcolor{gray!33!white}4.0 & \cellcolor{gray!33!white}4.0 & \cellcolor{gray!0!white}0.2 & \cellcolor{orange!91!white}9.889 \\
6 & \cellcolor{gray!33!white}4.0 & \cellcolor{gray!33!white}0.4 & \cellcolor{orange!34!white}6.351 & \cellcolor{gray!33!white}4.0 & \cellcolor{gray!33!white}0.4 & \cellcolor{orange!27!white}4.867 & \cellcolor{gray!33!white}4.0 & \cellcolor{gray!33!white}4.0 & \cellcolor{gray!33!white}0.4 & \cellcolor{orange!1!white}6.646 \\
7 & \cellcolor{gray!33!white}4.0 & \cellcolor{gray!66!white}0.6 & \cellcolor{orange!25!white}6.204 & \cellcolor{gray!33!white}4.0 & \cellcolor{gray!66!white}0.6 & \cellcolor{orange!31!white}5.036 & \cellcolor{gray!33!white}4.0 & \cellcolor{gray!33!white}4.0 & \cellcolor{gray!66!white}0.6 & \cellcolor{orange!0!white}6.591 \\
8 & \cellcolor{gray!33!white}4.0 & \cellcolor{gray!99!white}0.8 & \cellcolor{orange!8!white}5.934 & \cellcolor{gray!33!white}4.0 & \cellcolor{gray!99!white}0.8 & \cellcolor{orange!71!white}6.451 & \cellcolor{gray!33!white}4.0 & \cellcolor{gray!33!white}4.0 & \cellcolor{gray!99!white}0.8 & \cellcolor{orange!71!white}9.172 \\
9 & \cellcolor{gray!66!white}6.0 & \cellcolor{gray!0!white}0.2 & \cellcolor{orange!38!white}6.419 & \cellcolor{gray!66!white}5.0 & \cellcolor{gray!0!white}0.2 & \cellcolor{orange!77!white}6.676 & \cellcolor{gray!66!white}6.0 & \cellcolor{gray!66!white}5.0 & \cellcolor{gray!0!white}0.2 & \cellcolor{orange!93!white}9.978 \\
10 & \cellcolor{gray!66!white}6.0 & \cellcolor{gray!33!white}0.4 & \cellcolor{orange!24!white}6.185 & \cellcolor{gray!66!white}5.0 & \cellcolor{gray!33!white}0.4 & \cellcolor{orange!21!white}4.657 & \cellcolor{gray!66!white}6.0 & \cellcolor{gray!66!white}5.0 & \cellcolor{gray!33!white}0.4 & \cellcolor{orange!4!white}6.763 \\
11 & \cellcolor{gray!66!white}6.0 & \cellcolor{gray!66!white}0.6 & \cellcolor{orange!9!white}5.953 & \cellcolor{gray!66!white}5.0 & \cellcolor{gray!66!white}0.6 & \cellcolor{orange!19!white}4.593 & \cellcolor{gray!66!white}6.0 & \cellcolor{gray!66!white}5.0 & \cellcolor{gray!66!white}0.6 & \cellcolor{orange!9!white}6.946 \\
12 & \cellcolor{gray!66!white}6.0 & \cellcolor{gray!99!white}0.8 & \cellcolor{orange!10!white}5.969 & \cellcolor{gray!66!white}5.0 & \cellcolor{gray!99!white}0.8 & \cellcolor{orange!73!white}6.539 & \cellcolor{gray!66!white}6.0 & \cellcolor{gray!66!white}5.0 & \cellcolor{gray!99!white}0.8 & \cellcolor{orange!70!white}9.150 \\
13 & \cellcolor{gray!99!white}8.0 & \cellcolor{gray!0!white}0.2 & \cellcolor{orange!30!white}6.285 & \cellcolor{gray!99!white}6.0 & \cellcolor{gray!0!white}0.2 & \cellcolor{orange!81!white}6.803 & \cellcolor{gray!99!white}8.0 & \cellcolor{gray!99!white}6.0 & \cellcolor{gray!0!white}0.2 & \cellcolor{orange!99!white}10.166 \\
14 & \cellcolor{gray!99!white}8.0 & \cellcolor{gray!33!white}0.4 & \cellcolor{orange!8!white}5.928 & \cellcolor{gray!99!white}6.0 & \cellcolor{gray!33!white}0.4 & \cellcolor{orange!23!white}4.758 & \cellcolor{gray!99!white}8.0 & \cellcolor{gray!99!white}6.0 & \cellcolor{gray!33!white}0.4 & \cellcolor{orange!21!white}7.367 \\
15 & \cellcolor{gray!99!white}8.0 & \cellcolor{gray!66!white}0.6 & \cellcolor{orange!3!white}5.852 & \cellcolor{gray!99!white}6.0 & \cellcolor{gray!66!white}0.6 & \cellcolor{orange!0!white}3.923 & \cellcolor{gray!99!white}8.0 & \cellcolor{gray!99!white}6.0 & \cellcolor{gray!66!white}0.6 & \cellcolor{orange!8!white}6.924 \\
16 & \cellcolor{gray!99!white}8.0 & \cellcolor{gray!99!white}0.8 & \cellcolor{orange!4!white}5.877 & \cellcolor{gray!99!white}6.0 & \cellcolor{gray!99!white}0.8 & \cellcolor{orange!82!white}6.850 & \cellcolor{gray!99!white}8.0 & \cellcolor{gray!99!white}6.0 & \cellcolor{gray!99!white}0.8 & \cellcolor{orange!71!white}9.164 \\
\bottomrule
\end{tabular}%
}
\caption{Tabular result of action parameter with \textcolor{orange}{orange} hue heatmap for each interaction (better performance indicated by lighter color).}
\label{tab:action_config}
\end{table*}

\begin{table*}[!thb]
\centering
\resizebox{\textwidth}{!}{%
\begin{tabular}{c c ccc ccc ccc}
\toprule
 & & \multicolumn{3}{c}{Soft Objects} & \multicolumn{3}{c}{Hard Objects} & \multicolumn{3}{c}{Heterogeneous Objects} \\
\cmidrule(lr){3-5} \cmidrule(lr){6-8} \cmidrule(lr){9-11}
 & & Pressing & Precession & Sliding & Pressing & Precession & Sliding & Pressing & Precession & Sliding \\
\midrule
\multirow{5}{*}{\rotatebox{90}{Soft e-Skin}} 
& Spatial Freq. & \cellcolor{red!26!white}0.022 & \cellcolor{red!30!white}0.025 & \cellcolor{red!39!white}0.031 & \cellcolor{green!42!white}0.066 & \cellcolor{green!47!white}0.072 & \cellcolor{green!55!white}0.082 & \cellcolor{blue!4!white}0.055 & \cellcolor{blue!4!white}0.055 & \cellcolor{blue!5!white}0.070 \\
& Amplitude & \cellcolor{red!4!white}0.008 & \cellcolor{red!3!white}0.007 & \cellcolor{red!12!white}0.013 & \cellcolor{green!9!white}0.027 & \cellcolor{green!7!white}0.024 & \cellcolor{green!17!white}0.036 & \cellcolor{blue!0!white}0.009 & \cellcolor{blue!0!white}0.007 & \cellcolor{blue!0!white}0.010 \\
 & Stiffness & \cellcolor{red!41!white}0.032 & \cellcolor{red!64!white}0.047 & \cellcolor{red!72!white}0.052 & \cellcolor{green!80!white}0.112 & \cellcolor{green!77!white}0.108 & \cellcolor{green!99!white}0.134 & \cellcolor{blue!1!white}0.023 & \cellcolor{blue!1!white}0.030 & \cellcolor{blue!5!white}0.070 \\
& Heterogeneity & \cellcolor{red!18!white}0.017 & \cellcolor{red!16!white}0.016 & \cellcolor{red!58!white}0.043 & \cellcolor{green!1!white}0.017 & \cellcolor{green!8!white}0.025 & \cellcolor{green!14!white}0.032 & \cellcolor{blue!43!white}0.516 & \cellcolor{blue!31!white}0.376 & \cellcolor{blue!53!white}0.629 \\
\cmidrule(lr){2-11}
& Overall & \cellcolor{red!18!white}0.017 & \cellcolor{red!27!white}0.023 & \cellcolor{red!46!white}0.035 & \cellcolor{green!30!white}0.052 & \cellcolor{green!34!white}0.057 & \cellcolor{green!45!white}0.070 & \cellcolor{blue!10!white}0.130 & \cellcolor{blue!8!white}0.105 & \cellcolor{blue!15!white}0.191 \\
\midrule
\multirow{5}{*}{\rotatebox{90}{Hard e-Skin}} & Spatial Freq. & \cellcolor{red!24!white}0.021 & \cellcolor{red!13!white}0.014 & \cellcolor{red!53!white}0.040 & \cellcolor{green!30!white}0.052 & \cellcolor{green!26!white}0.047 & \cellcolor{green!67!white}0.096 & \cellcolor{blue!3!white}0.045 & \cellcolor{blue!3!white}0.048 & \cellcolor{blue!6!white}0.080 \\
& Amplitude & \cellcolor{red!3!white}0.007 & \cellcolor{red!1!white}0.006 & \cellcolor{red!10!white}0.012 & \cellcolor{green!5!white}0.022 & \cellcolor{green!2!white}0.018 & \cellcolor{green!21!white}0.041 & \cellcolor{blue!0!white}0.008 & \cellcolor{blue!0!white}0.008 & \cellcolor{blue!0!white}0.017 \\
& Stiffness & \cellcolor{red!56!white}0.042 & \cellcolor{red!59!white}0.044 & \cellcolor{red!99!white}0.070 & \cellcolor{green!60!white}0.088 & \cellcolor{green!57!white}0.084 & \cellcolor{green!91!white}0.125 & \cellcolor{blue!2!white}0.031 & \cellcolor{blue!3!white}0.047 & \cellcolor{blue!7!white}0.094 \\
& Heterogeneity & \cellcolor{red!75!white}0.054 & \cellcolor{red!72!white}0.052 & \cellcolor{red!35!white}0.028 & \cellcolor{green!0!white}0.015 & \cellcolor{green!9!white}0.027 & \cellcolor{green!11!white}0.029 & \cellcolor{blue!46!white}0.543 & \cellcolor{blue!58!white}0.685 & \cellcolor{blue!99!white}1.172 \\
\cmidrule(lr){2-11}
& Overall & \cellcolor{red!41!white}0.032 & \cellcolor{red!36!white}0.029 & \cellcolor{red!49!white}0.037 & \cellcolor{green!20!white}0.040 & \cellcolor{green!22!white}0.042 & \cellcolor{green!47!white}0.072 & \cellcolor{blue!10!white}0.130 & \cellcolor{blue!14!white}0.176 & \cellcolor{blue!28!white}0.339\\ 
\bottomrule
\end{tabular}%
}
\caption{Tabular result with NMSE values over the entire interaction sequence (lower is better) analyzing whether there is inherent advantage of the properties of object and mechanical properties of e-Skin. The overall value is computed by taking the mean over all the dimensions of softness at the last-time step, post convergence.}
\label{tab:environmentshape}
\end{table*}

\subsection{Analysis on Stiffness of e-Skin}
\label{subsec:reg3}
An analysis of the influence of the e-Skin’s mechanical stiffness on perception was conducted using the MULTI-L configuration (see supplementary material, Fig\,3). The findings indicate that hard skin (DragonSkin-30) consistently yields better performance in estimating spatial frequency, amplitude, and bulk stiffness, whereas softer skin (Ecoflex-00-31) achieves improved accuracy when interacting with complex heterogeneous objects, particularly during sliding interactions. However, because the distribution of soft and hard objects in the dataset was not uniform, and to further explore whether perceptual performance is influenced by the mechanical similarity between the skin and the object ~\cite{costi2022environment}, we grouped the objects into three categories: soft (Objects {0,1,3,4,9,10,12,13,18,19,21,22}), heterogeneous (Objects {27–31}), and hard. NMSE errors were then recomputed for each group over the interaction sequence and are reported in Table~\ref{tab:environmentshape}.

This object-specific analysis reveals that soft e-Skin offers a perceptual advantage when interacting with soft objects, though this benefit is primarily observed during pressing interactions and is most prominent in the estimation of surface heterogeneity and bulk stiffness. This advantage is particularly evident during sliding and pressing interactions. However, this difference is diminished for precession interaction were stable and stronger contact was established between the e-Skin and the soft objects. Conversely, hard e-Skin performs more reliably across hard objects, maintaining consistent accuracy across interaction types. Notably, soft skin is particularly well-suited for perceiving the properties of heterogeneous materials with spatially varying viscoelasticity, as well as for surface-level interactions such as sliding. Overall, the results highlight that the perceptual effectiveness of a given e-Skin stiffness depends both on object properties and on the interaction type, suggesting the opportunities offered by co-designing embodiment of tactile sensing together with exploration strategies.

\section{Discussion}
\label{sec:discuss}
This paper introduces a principled framework for investigating how variations in interaction strategies and sensor embodiment influence tactile perception of soft and rigid objects. It presented three core contributions: (i) the development of a modular, multi‑modal \textit{e‑Skin} tactile sensor with tunable mechanical compliance, capable of capturing both spatial force and vibration signals; (ii) the design of a foundational dataset of \textit{wave objects} that systematically varies in viscoelasticity, surface texture, and spatial heterogeneity, representing the diversity of natural environments; and (iii) the \textit{Latent Filter}, an unsupervised, action‑conditioned deep state‑space model that encodes interaction dynamics into a structured latent manifold for causal inference of object properties.

% Discussion on multi-modality & fusion
Our findings from diverse interaction scenarios highlight several important aspects of tactile perception. High‑dimensional, multi‑modal sensing proved essential for generalization, as different modalities dominated in different interaction contexts. The tactile perception through sensors embodiment and the latent filter. As in humans \cite{Hillis2002}, the fusion of multiple modalities with the same sense offers benefits relative to each of the separate modalities. This inclusion of two FSR layers was particularly beneficial for estimating stiffness and heterogeneity, since differential normal forces provide a proxy for shear forces and local skin deformation. Although accelerometer‑only embeddings were less discriminative overall, they captured valuable transient information during surface‑level interactions such as sliding—information that the FSR layers alone could not provide. The modular, layered design of the e-Skin opens new avenues for exploring fine-grained mechanical effects such as the role of artificial fingerprints, and enable robotic system with enhanced tactile embodiment through both mechanical versatility and rich sensing capabilities.

Importantly, sensory processing with late‑fusion of different modalities consistently outperformed early fusion and uni-modal approaches. This suggests that considering separate encoding pathways prior to integration can prevent representations mismatch that may arise when combining signals with different temporal and spatial statistics. These results indicate the need for modality‑aware tactile encoders that can explicitly disentangle force‑ and vibration‑based features, which is a promising direction for achieving robust and interpretable perception.

% Discussion on interaction
The motion interaction strategy also emerged as a critical factor shaping perception. By incorporating temporal dynamics, the latent filtering approach with regression technique in Section~\ref{subsec:reg1}, \ref{subsec:reg3}, \ref{subsec:reg3} achieved more interpretable and sensitive representations than static latent space analyses such as UMAP or Euclidean distance metrics presented in Section~\ref{subsec:static}. Motion with moderate indentation frequencies ($\sim0.6$\,Hz) and sufficient contact depth consistently resulted in low NMSE across properties, with sliding and precession interactions benefiting most from parameter tuning. In contrast, pressing was less sensitive to parameter variation, mainly affecting convergence speed rather than final estimation accuracy. These observations are consistent with findings from human haptic exploration, where perception is actively modulated by task‑specific exploratory procedures~\cite{ lederman1993extracting}. The results therefore suggest that robotic palpation strategies should be adaptively selected according to the property of interest and interaction strategy.

% Discussion on mechanical properties
The mechanical stiffness of the e‑Skin further influenced perceptual performance in a property‑ and interaction‑specific manner. A stiffer skin (with DragonSkin) led to more accurate bulk property estimation due to improved force transmission, whereas a softer skin (with Ecoflex) enhanced the perception of heterogeneous or compliant objects, particularly in sliding interactions where surface‑level features dominate. This suggests that the embodiment of tactile sensing could be co‑optimized with interaction strategies rather than treated as a fixed design choice. In addition, the latent filter framework proved effective in extracting causal, interpretable representations of object properties by leveraging action‑conditioned temporal dynamics. By explicitly modeling both directly and indirectly observable latent components, the approach bridges the gap between discriminative deep learning methods and analytical models, thereby supporting generalizable perception across diverse soft objects.  This framework holds significant potential for soft robotic applications, including medical palpation, food manipulation, and the handling of delicate or deformable objects.

In future, we will focus on closed-loop active perception, where the robot autonomously selects optimal interaction strategies based on current uncertainty, and on developing programmable e-Skins with real-time tunable compliance. Additionally, integrating the latent filter with reinforcement learning or model-based control could lead to perception-aware policies that exploit the inferred physical properties for more dexterous and reliable manipulation. In conclusion, this study strengthens the importance of jointly optimizing sensor design and interaction strategies, highlighting that tactile perception cannot be decoupled from action or embodiment.

\section{Methods}
\label{sec:methods}
In this section, we describe the controlled palpation strategies used for tactile interaction and the data processing pipeline employed to prepare the observations for analysis and model training.

\subsection{Palpation Interaction}
\label{subsec:interaction}
The e-Skin was mounted on the end-effector of a 7-DOF robotic arm. To generate a rich dataset of interactions, we implemented three robotic palpation primitives (Fig.\,\ref{fig:methodsfig}c), combining translational and rotational motions to probe both surface and bulk(internal) object properties. \textit{Pressing} consists of a sinusoidal motion normal to the object surface. \textit{Precession} with a rotational movement involving roll ($\alpha$) and pitch ($\beta$), providing contact with multiple surface regions for probing complex geometries with concave and convex features. \textit{Sliding} emulated natural human texture exploration, capturing surface roughness, structural details, and friction. These movements are defined as:
\begin{align}
    pressing: z(t) &= A_z \cos\left(2\pi \omega_z  t + \pi\right) \nonumber \\
    precession: \alpha(t) &= A_\alpha \sin\left(2\pi \omega_\alpha t + \pi/2\right) \nonumber  \\
    \beta(t) &= A_\beta \sin \left(2\pi \omega_\beta t\right) \\
     sliding: x(t) &= A_x \sin\left(2\pi \omega_x  t\right) \nonumber  \\
     y(t) &= A_y \sin\left(2\pi \omega_y  t\right) \nonumber  \\
     \gamma(t) &= A_\gamma \sin\left( 2\pi \omega_\gamma t\right) \nonumber 
\end{align}
where $t$ represents the time, and the amplitude parameters $A_{(\cdot)}$ and frequency parameters $\omega_{(\cdot)}$ are specific to each movement type. The Kinova robotic arm was programmed within a ROS2 \cite{ros2} environment to execute these interaction trajectories with precise temporal control. Tactile data from the e‑Skin and proprioceptive data from the robot (end‑effector pose, joint positions, and applied forces) were synchronously recorded using a custom data acquisition pipeline. A lightweight Protocol Buffers (Protobuf) \cite{protobuf} schema was implemented to efficiently serialize multi‑modal sensor streams, with each message containing a high‑resolution timestamp. This ensured time‑aligned logging of tactile observations and robot proprioception, providing accurate correspondence between sensor events and executed interaction trajectories for subsequent model training and analysis.

\subsection{Data processing \& training}
\label{subsec:dataprocessing}
Two transformations were applied to address the challenges of training on long sequences. For the accelerometer data (7000$\times$4$\times$4$\times$3), the normalized magnitude was computed and downsampled to 600\,Hz, resulting in a (6000$\times$4$\times$4) time series. A log-mel spectrogram \cite{mcfee2015librosa} was then calculated using a window size of 800, hop length of 20, 128 FFT points, and 49 mel bins, yielding a (300$\times$49$\times$4$\times$4) representation. This was reshaped into (300$\times$28$\times$28) time-series sequences, preserving both spatial structure and temporal dynamics.

\begin{figure}[!th]
    \centering
    \includegraphics[width=\textwidth]{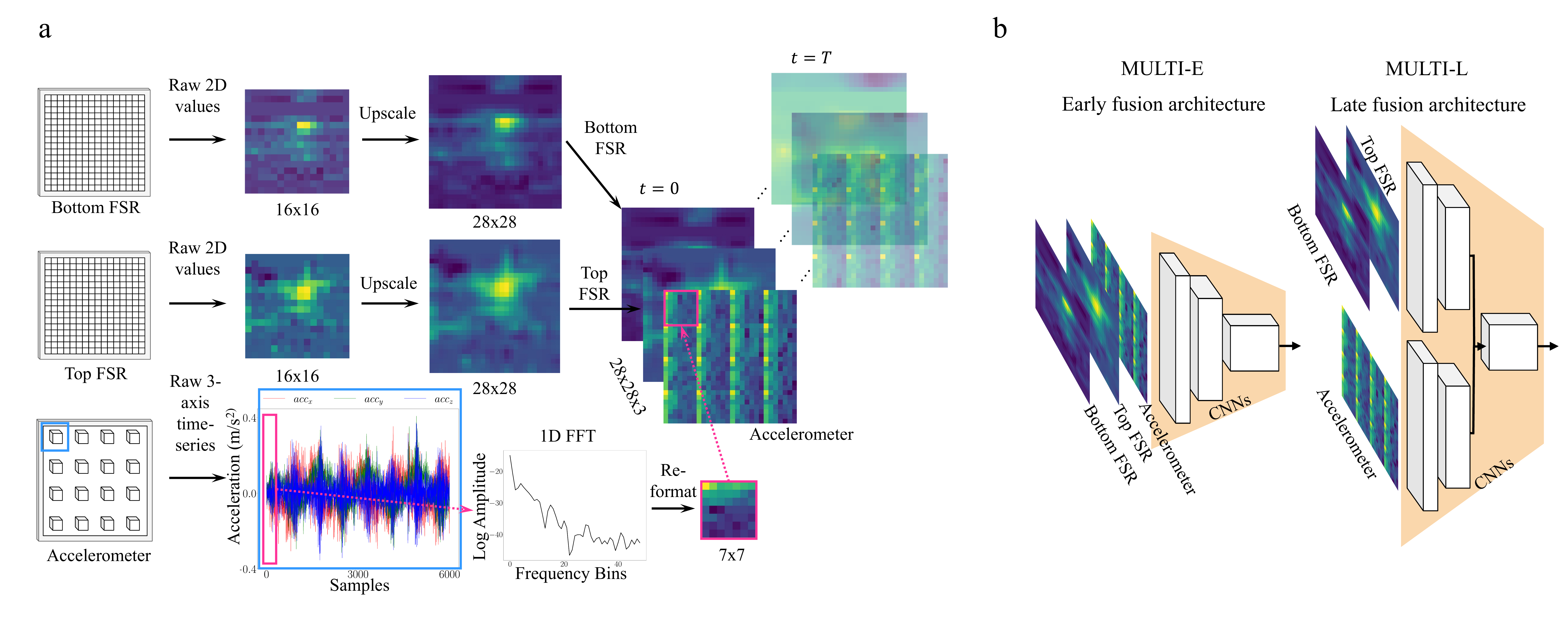}
    \caption{a) Visualization of the preprocessing step of tactile observations b) Encoder architecture overview of two fusion techniques to combine accelerometer and FSR observations within the \textit{latent filter}. Multi-E represents early fusion approach, while Multi-L denotes the late fusion technique.}
    \label{fig:multifusionarc}
\end{figure}

For the FSR data (300$\times$16$\times$16$\times$2), mean filtering was applied and the data was downsampled to 30\,Hz. To enable evaluation of fusion strategies that integrate spatial (force, shear) and spectral (vibration) information, the FSR signals were spatially rescaled to (300$\times$28$\times$28$\times$2) using the interpolation function from \cite{paszke2019pytorch}. This consistent dimensionality across the modalities facilitated comparison between early and late fusion techniques and supported subsequent ablation studies. A graphical overview of the preprocessing pipeline is shown in Figure \ref{fig:multifusionarc}a.

Twenty-five percent of the dataset was reserved for testing. The \textit{Latent Filter} model was implemented in PyTorch and trained using the Adam optimizer with a learning rate of $10^{-5}$. The training process employed an annealing strategy \cite{karl2017deep} to gradually increase the regularization weight in the ELBO loss, encouraging stable and effective disentanglement in the latent space. Careful hyperparameter tuning was performed for the learning rate, latent dimensionality, and the temperature parameter of the annealing term. The latent dimension was $n_s = 32$ (comprising $n_y = 16$ and $n_z = 16$). Model training was performed on a workstation running Ubuntu\,20.04\,LTS, equipped with an Intel Xeon 5222 CPU, 32GB RAM, and an NVIDIA RTX\,A4000 GPU (16\,GB VRAM). Each training run converged within approximately 12 hours for 500 epochs using a batch size of 48. In total 24 models - $8$ (combination of modality and e‑Skin types) $\times$ $3$ (interaction primitives) were trained.

\section{Data Availability}
The full dataset generated and/or analyzed during this study is not publicly available due to its large size ($\sim$200 GB), but it is available from authors upon reasonable request.  A representative subset is publicly available at: 
\url{https://drive.google.com/file/d/1kAkIVdUE4aZ14MwaYkYG_7kUjKIVOQVl/view?usp=sharing} for evaluation.

\section{Code Availability}
The code supporting this study is publicly available on the GitHub repository \texttt{EmbodiedTactileSoftObjects}, which can be accessed at:
\url{https://github.com/anirvan95/EmbodiedTactileSoftObjects.git}.

\section{Acknowledgments}
The authors thank Dr. Mohsen Kaboli for discussions on this project. The work was supported in part by the EC H2020 grants INTUITIVE (ITN 861166) and PH-CODING (FETOPEN 829186). Anirvan Dutta was also associated with BMW Group AG, M\"unchen Germany, during this work. 

\section{Author contributions}
All authors contributed to the conceptualization, experimental design, and analysis of the research. A. Dutta implemented and developed the latent filtering framework, A. Devillard developed the e‑Skin, and Z. Zhang conducted the experimental data collection using robotic hardware. X. Cheng and E. Burdet provided supervision and guidance throughout the project. All authors were involved in the writing and revision of the manuscript.

\section{Competing Interests}
All authors declare no financial competing interests. 

\bibliography{sn-bibliography}

\section{Supplementary Materials}

\subsection{Derivation of Latent Filter}
\label{subsec:latentfilterder}

We assume a generative model with an underlying latent dynamical system with  
\begin{align}
    p(\mathbf{o}_{1:T}|\mathbf{a}_{1:T}) =  \int \! \! p(\mathbf{o}_{1:T}&|\mathbf{s}_{1:T}, \mathbf{a}_{1:T}) \, p(\mathbf{s}_{1:T}|\mathbf{a}_{1:T}) \, d\mathbf{s}_{1:T} \nonumber  \\
    = \int \prod_{t=1}^{T}p(\mathbf{o}_{t}|\mathbf{o}_{1:t-1}, \mathbf{s}_{t}, \mathbf{s}_{1:t-1}, \mathbf{a}_{1:t})&\,p(\mathbf{s}_{t}|\mathbf{o}_{1:t-1}, \mathbf{s}_{t-1}, \mathbf{s}_{1:t-2}, \mathbf{a}_{t}, \mathbf{a}_{1:t-1}) \, d\mathbf{s}_{1:t} 
    \label{eq:generativemodelgen}
\end{align}
\noindent Eq.\,\ref{eq:generativemodelgen} becomes computationally expensive as the number of time steps increases, due to the growing complexity of conditional dependencies. To address this, we adopt the first-order Markov assumption, which states that (i) the latent state at time $t$ depends only on the immediate past state $\mathbf{s}_{t-1}$ (rather than on the full history $\mathbf{s}_{1:t-1}$) and that (ii) the observations at each time step are fully determined by the current state. This assumption enables a more compact factorization of the joint distribution and facilitates tractable learning and inference over long temporal sequences
\begin{align}
    p(\mathbf{o}_{t}|\cancel{\mathbf{o}_{1:t-1}}, \mathbf{s}_{t}, \cancel{\mathbf{s}_{1:t-1}}, \cancel{\mathbf{a}_{1:t}}) &= p(\mathbf{o}_{t}|\mathbf{s}_{t}) \\
    p(\mathbf{s}_{t}|\cancel{\mathbf{o}_{1:t-1}}, \mathbf{s}_{t-1}, \cancel{\mathbf{s}_{1:t-2}}, \mathbf{a}_{t}, \cancel{\mathbf{a}_{1:t-1}}) &= p(\mathbf{s}_{t}|\mathbf{s}_{t-1}, \mathbf{a}_{t}) \nonumber 
\end{align}
resulting in the simplified generative model: 
\begin{align}
    p(\mathbf{o}_{1:T}|\mathbf{a}_{1:T}) = \,p(\mathbf{o}_1|\mathbf{s}_{1},\mathbf{a}_{1} )\, p(\mathbf{s}_1)
 \!   \int \! \prod_{t=2}^T p(\mathbf{o}_t|\mathbf{s}_t) \, p(\mathbf{s}_t|\mathbf{s}_{t-1}, \mathbf{a}_t) \, d\mathbf{s}_t\,.
    \label{eq:generativemodelmarkov}
\end{align}
With the proposed factorization results in the following generative model and factorization from Eq.\ref{eq:generativemodelmarkov}
\begin{align}
   p(\mathbf{o}_{1:T}|\mathbf{a}_{1:T}) = \iint \prod_{t=1}^T \, p(\mathbf{o}_t|\mathbf{z}_t) \, p(\mathbf{z}_t|\mathbf{z}_{t-1},\mathbf{y}_t, \mathbf{a}_t)\, p(\mathbf{y}_t|\mathbf{y}_{t-1}, \mathbf{a}_{t-1}) \, d\mathbf{y}_{t-1}d\mathbf{z}_{t-1}
    \label{eq:generativemodelprop}
\end{align}
To compute the observation likelihood, we introduce variational distribution $q_\theta(\mathbf{z}_{1:T}, \mathbf{y}_{1:T}) \sim p(\mathbf{z}_{1:T}, \mathbf{y}_{1:T}|\mathbf{o}_{1:T}, \mathbf{a}_{1:T})$. The \textit{Evidence Lower Bound Objective} (ELBO) for the generative model in Eq. \ref{eq:generativemodelprop} is formulated from the \textsc{KL} Divergence inequality:
\begin{align}
    D_{\textsc{KL}} = - \! \iint \! q_\theta(\mathbf{z}_{1:T}, \mathbf{y}_{1:T}|\mathbf{o}_{1:T}, \mathbf{a}_{1:T})
    \log \left[\frac{p(\mathbf{z}_{1:T}, \mathbf{y}_{1:T}|\mathbf{o}_{1:T}, \mathbf{a}_{1:T}}{q_\theta(\mathbf{z}_{1:T}, \mathbf{y}_{1:T}|\mathbf{o}_{1:T}, \mathbf{a}_{1:T}}\right] d\mathbf{z}_{1:T} \, d\mathbf{y}_{1:T}\geq 0
\end{align}
resulting in the following objective function
\begin{align}
    \log p(\mathbf{o}_{1:T}|\mathbf{a}_{1:T}) \geq \mathbb{E}_{q_\theta(.)} \! \left[ \log p(\mathbf{o}_{1:T}|\mathbf{z}_{1:T}, \mathbf{y}_{1:T}, \mathbf{a}_{1:T}) \right] \nonumber \\
    -  \mathbb{E}_{q_\theta(.)} \! \left[ \log \! \left(\frac{q_\theta(\mathbf{z}_{1:T}, \mathbf{y}_{1:T}|\mathbf{o}_{1:T}, \mathbf{a}_{1:T})}{p(\mathbf{z}_{1:T}, \mathbf{y}_{1:T}|\mathbf{a}_{1:T})}\right) \! \right] \, .
\end{align}
Re-introducing Markov assumption and simplification presented in the generative model in the inference into the regularization term of the ELBO
\begin{align}
    q_\theta(\mathbf{z}_{1:T}, \mathbf{y}_{1:T}  |\mathbf{o}_{1:T}, \mathbf{a}_{1:T}) &= 
    q_{\theta}(\mathbf{z}_1|\mathbf{y}_1) \, q_{\phi}(\mathbf{y}_1) \prod_{t=2}^{T} \, q_\theta(\mathbf{z}_t, \mathbf{y}_t|\mathbf{z}_{t-1}, \mathbf{y}_{t-1}, \mathbf{o}_{1:t}, \mathbf{a}_{1:t}) \noindent \\
    q_\theta(\mathbf{z}_t, \mathbf{y}_t|\mathbf{z}_{t-1}, \mathbf{y}_{t-1}, \mathbf{o}_{1:t}, \mathbf{a}_{1:t}) &= \frac{p(\mathbf{o}_t|\mathbf{z}_t) \, q_\theta(\mathbf{z}_t|\mathbf{z}_{t-1}, \mathbf{y}_t, \mathbf{a}_t)q_\phi(\mathbf{y}_t|\mathbf{z}_{t-1}, \mathbf{y}_{t-1}, \mathbf{a}_{t-1})}{\iint p(\mathbf{o}_t|\mathbf{z}_t) \, p(\mathbf{z}_t|\mathbf{z}_{t-1}, \mathbf{y}_t, \mathbf{a}_t) \, d\mathbf{z}_{t-1}d\mathbf{y}_t} \nonumber \\
    \sim  \eta &\underbrace{q_{\psi}(\mathbf{z}_t|\mathbf{o}_t) \, q_\theta(\mathbf{z}_t|\mathbf{z}_{t-1}, \mathbf{y}_t, \mathbf{a}_t)}_{q^{filt}(\mathbf{z}_t)}q_\phi(\mathbf{y}_t|\mathbf{z}_{t-1}, \mathbf{y}_{t-1}, \mathbf{a}_{t-1})
\end{align}
where the denominator is the normalizer, results in the simplified ELBO:
\begin{align}
    \mathcal{F}_{\textsc{ELBO}}(\theta, \phi)  = \,\mathbb{E}_{q_\phi(.)}\!\left[ \sum_{t=1}^{T} \log p(\mathbf{o}_t|\mathbf{z}_t)\right] 
    - &\sum_{t=2}^{T}{\textsc{KL}}[ q^{filt}(\mathbf{z}_t) || p(\mathbf{z}_{t}|\mathbf{z}_{t-1}, \mathbf{y}_{t}, \mathbf{a}_{t}) \nonumber \\
    -  \sum_{t=2}^{T}{\textsc{KL}}[ &q_\phi(\mathbf{y}_t|.)||p(\mathbf{y}_t| \mathbf{a}_{t})]
\end{align}

\subsection{Additional Results}
\label{subsec:addresults}
\begin{figure}[!ht]
    \centering
    \begin{subfigure}[b]{0.49\textwidth}
        \centering
        \includegraphics[width=\textwidth]{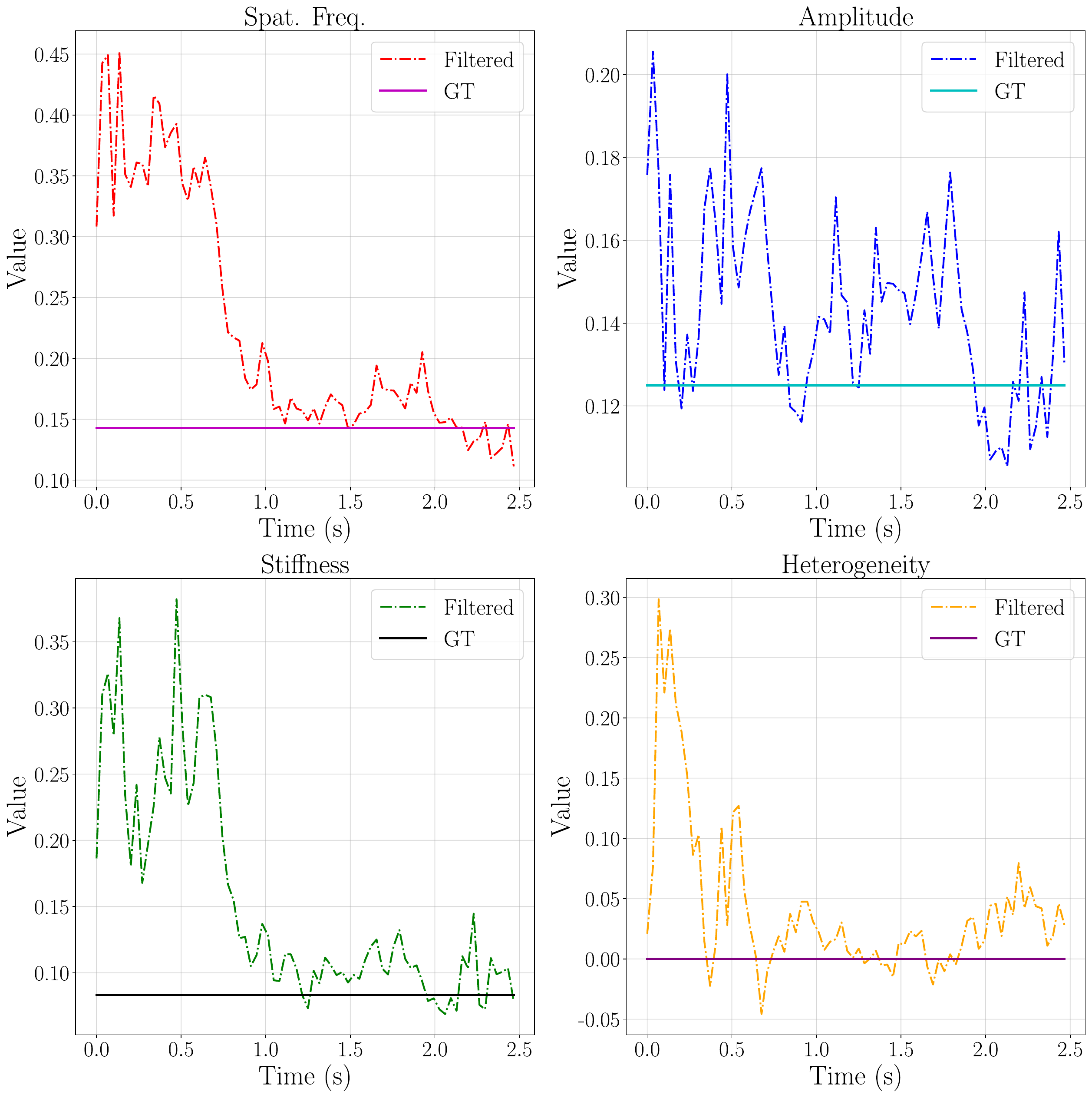}
        \caption{Filtering in action object 0 action parameter 1}
        \label{fig:filtobj1}
    \end{subfigure}
    \hfill
    \begin{subfigure}[b]{0.49\textwidth}
        \centering
        \includegraphics[width=\textwidth]{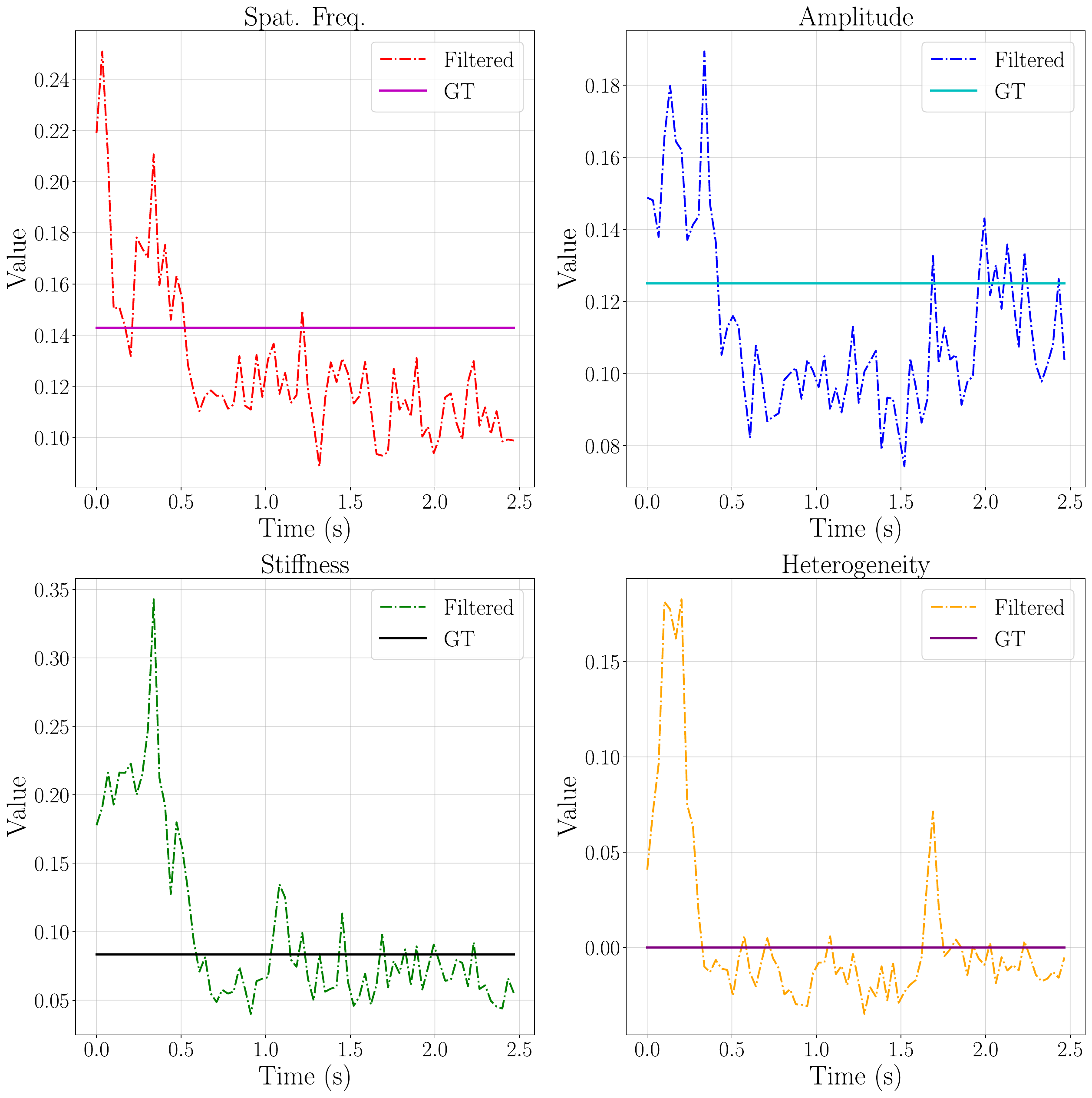}
        \caption{Filtering in action object 0 action parameter 15}
        \label{fig:filtobj2}
    \end{subfigure}
    \caption{Filtering results for object 0 with different action parameters}
    \label{fig:combinedfiltering}
\end{figure}

\begin{figure}[!ht]
    \centering
    \includegraphics[width=0.85\textwidth]{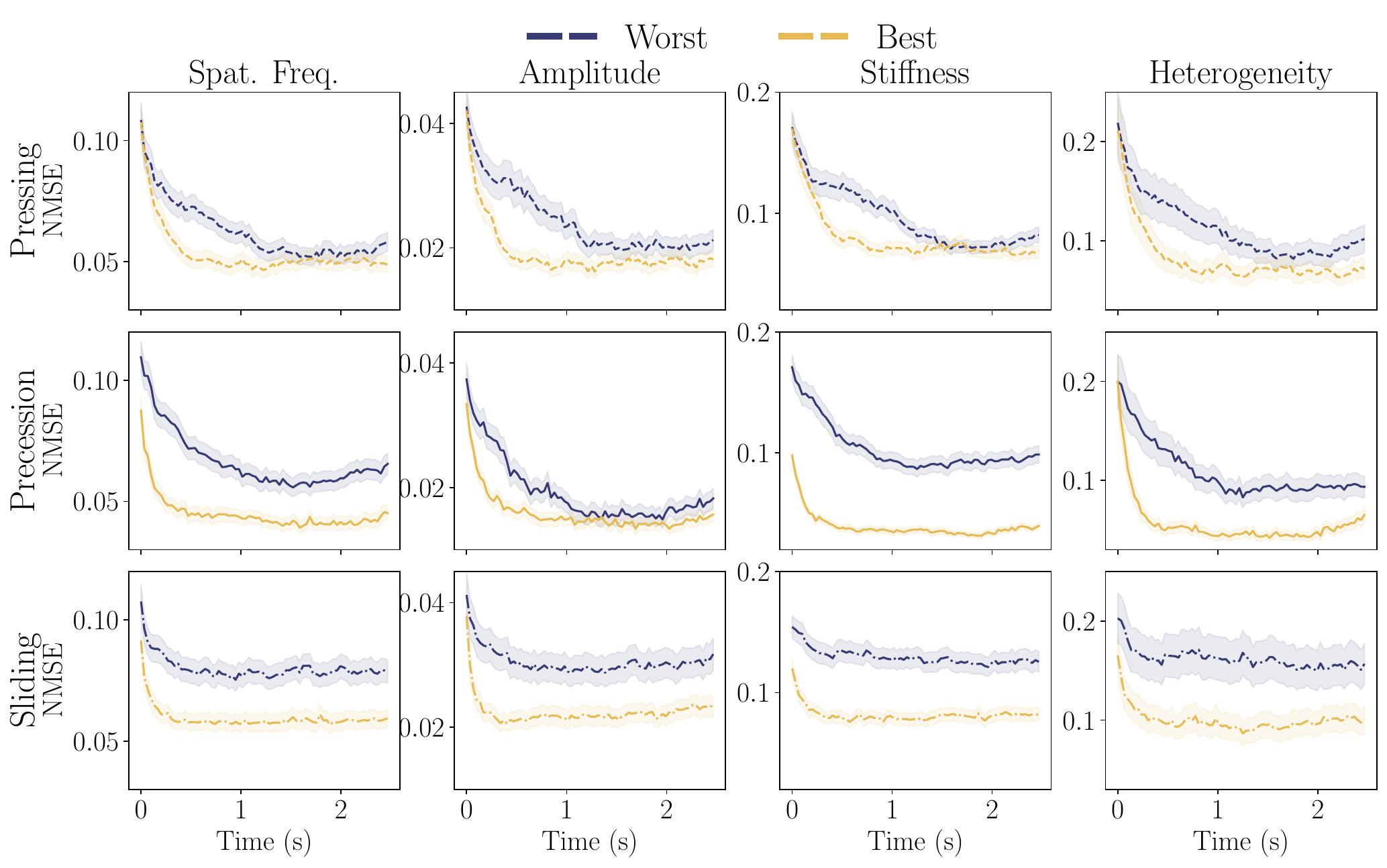}
    \caption{Comparison of different action parameters for MULTI-L (Soft) approach. The best and worst performing interaction parameters are selected to avoid visual and interpretive clutter.}
    \label{fig:action}
\end{figure}
\begin{figure}[!ht]
    \centering
    \includegraphics[width=0.85\textwidth]{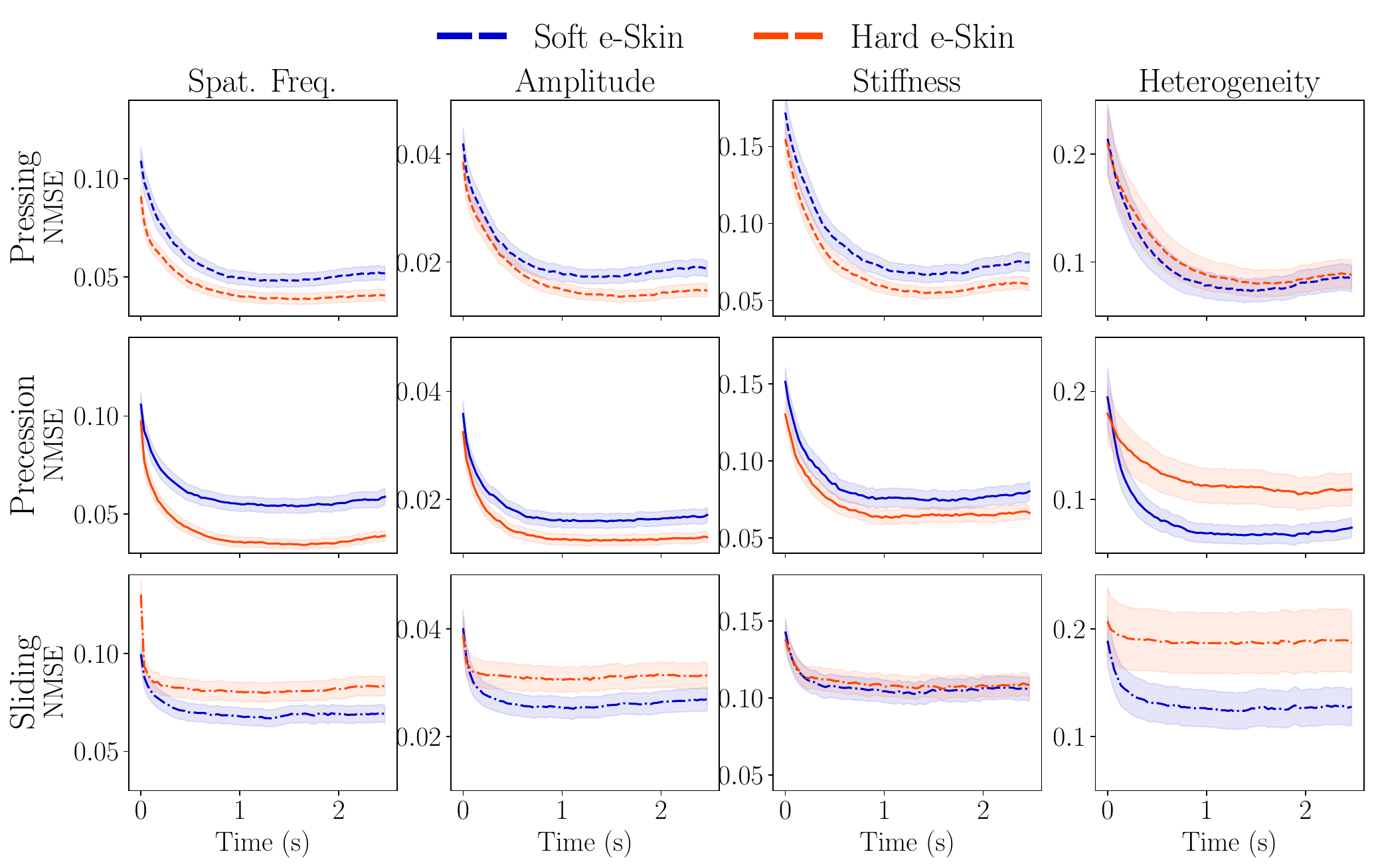}
    \caption{Comparison between hard and soft skin types using MULTI-L (multi-modal late fusion) approach.}
    \label{fig:stiffness}
\end{figure}

\end{document}